\documentclass[3p,times,onecolumn]{elsarticle}

\usepackage{graphicx}
\usepackage{subfigure}

\usepackage{amssymb}
\usepackage{amsmath}

\usepackage{url}
\usepackage[dvipsnames]{xcolor}

\makeatletter
\newif\if@restonecol
\makeatother

\usepackage{bm}
\usepackage{multirow}
\usepackage{mathrsfs}

\usepackage[ruled,linesnumbered,lined]{algorithm2e}

\usepackage{boolexpr,pdftexcmds,trace}
\usepackage{array} 

\usepackage{pdflscape}

\usepackage[disable]{todonotes} 

\allowdisplaybreaks

\makeatletter
\newcommand{\customlabel}[2]{%
\protected@write \@auxout {}{\string \newlabel {#1}{{#2}{}}}}
\makeatother

\def\clap#1{\hbox to 0pt{\hss#1\hss}}

\DeclareTextAccent{\myacc}{T1}{4}

{\end{cases}\right\rbrace}

\usepackage{fancyhdr}
\usepackage{multirow}
\usepackage{booktabs}

\usepackage{float}
\usepackage{pgffor}
\usepackage{morefloats}
\usepackage{xspace}

\usepackage{hyperref}

\usetikzlibrary{matrix}

\newcolumntype{C}[1]{>{\centering\let\newline\\\arraybackslash\hspace{0pt}}m{#1}}
\newcolumntype{L}[1]{>{\raggedright\let\newline\\\arraybackslash\hspace{0pt}}m{#1}}

\begin{document}

\begin{frontmatter}



\title{A New Vision of Collaborative Active Learning}
\author[IES]{Adrian Calma}
\author[IES]{Tobias Reitmaier}
\author[IES]{Bernhard Sick}
\author[DFKI]{Paul Lukowicz}
\author[RPI]{Mark Embrechts}

\address[IES]{Intelligent Embedded Systems Lab, University of Kassel, Kassel, Germany\\ 
(e-mail: $\{\text{adrian.calma,tobias.reitmaier,bsick}\}\text{@uni-kassel.de}$)}
\address[DFKI]{Embedded Intelligence, German Research Center for Artificial Intelligence, University of Kaiserslautern, Kaiserslautern, Germany\\ 
(e-mail: $\text{paul.lukowicz@dfki.de}$)}
\address[RPI]{Rensselaer Polytechnic Institute, Troy, New York, USA\\
(e-mail: $\text{embrem@rpi.edu}$)}


\begin{abstract}
Active learning (AL) is a paradigm where an active learner has to train a model (e.g., a classifier) which is in principle trained in a supervised way. In contrast to supervised learning, AL has to be done by means of a data set where a low fraction of samples is labeled or even with an initially unlabeled set of samples. To obtain labels for the unlabeled samples, the active learner has to ask an oracle (e.g., a human expert) for labels. In most cases, the goal is to maximize some metric or task performance and to minimize the number of queries at the same time. In this article, we first briefly discuss the state-of-the-art and own, preliminary work in the field of AL. Then, we propose the concept of collaborative active learning (CAL). With CAL, we will overcome some of the harsh limitations of current AL. In particular, we envision scenarios where the expert or the gold standard may be wrong for various reasons. There also might be several or even many experts with different expertise, the experts may label not only samples but also supply knowledge at a higher level such as rules, and we consider that the labeling costs depend on many conditions. Moreover, in a CAL process human experts may even profit by improving their own knowledge, too.
\end{abstract}


\end{frontmatter}


\section{Introduction}\label{sec:Introduction}


\todo{TR@all: Was ist mit weak learner (mechanical turk)? BS@TR: im Prinzip interessant aber wie und wo dar\"uber schreiben?}
Machine learning is based on sample data. Sometimes, these data are labeled and, thus, models to solve a certain problem (e.g., a classification or regression problem) can be built using targets assigned to input data of the model. In other cases, data are unlabeled (e.g., for clustering problems) or only partially labeled. Correspondingly, we distinguish the areas of supervised, unsupervised, and semi-supervised learning. In many application areas (e.g., industrial quality monitoring processes \cite{Sic98}, intrusion detection in computer networks \cite{HSS03}, speech recognition \cite{FHYA12}, or drug discovery \cite{MME14}) it is rather easy to collect unlabeled data, but quite difficult, time-consuming, or expensive to gather the corresponding targets. That is, labeling is in principal possible, but the costs may be enormous.

This article focuses on \textit{active learning (AL)}, a machine learning area that is sometimes considered to be a special case of semi-supervised learning (SSL) \cite{CSZ06}, but it can also be regarded as being closely related to SSL. AL starts with an initially unlabeled or very sparsely labeled set of samples and iteratively increases the labeled fraction of the training data set by ``asking the right questions''. These questions are then answered by humans (e.g., experts in an application domain), by simulation systems, by means of real experiments, etc., often modeled by an abstract ``oracle''. Basically, the ``idealized'' goal of AL is to obtain a model with (almost) the performance of a model trained with a fully labeled data set at (almost) the cost of an unlabeled data set. Thus, we also can say that SSL has to \textit{exploit} the knowledge contained in the unlabeled data, while AL has to \textit{explore} the knowledge contained in the unlabeled data.
Typically, the following assumptions are made (amongst others) in AL:
\begin{enumerate}
	\item The oracle labels single samples or sets of samples (called queries depending on the AL type, see below) presented by an active learner.
	\item The oracle is omniscient and omnipresent, i.e., it always delivers the correct answers and it is always available).
	\item The labeling costs for all samples are identical.
\end{enumerate}
This article develops a vision to overcome these limitations that are definitely not realistic for many applications. In particular, we assume that
\begin{enumerate}
	\item an expert may be wrong for various reasons, e.g., depending on her/his experience in the application domain (we still assume we have no ``malicious or deceptive experts'' that cheat or attack the active learner),
	\item there might be several or even many experts with different expertise (e.g., degree or kind of experience),
	\item the experts may label not only samples but also infer knowledge at a higher level such as rules (e.g., by assigning a conclusion to a presented premise), and
	\item the labeling costs depend on many conditions, e.g., whether samples or rules are labeled, on the location of samples in the input space of a model (i.e., making labeling more or less difficult), the degree of expertise of a human, etc.
\end{enumerate}
Moreover, there may be several tasks that have to be fulfilled at the same time (e.g., movies that are assessed regarding several criteria) and different kinds of information sources (e.g., human experts and simulation systems).

We envision \textit{collaborative active learning (CAL)} approaches where the above limitations no longer hold. Moreover, the humans involved in such a CAL process shall be offered the opportunity to profit by improving their own knowledge.

The field of AL recently has awoken the interest of many companies, such as Microsoft, IBM, Siemens, AT$\&$T, Mitsubishi, or Yahoo. Existing publications of those companies show that AL can be successfully utilized to solve a wide range problems, such as:
\begin{itemize}
	\item \textit{Text Classification:} Microsoft Research describes in \cite{PVS+10} an AL approach for providing personalized news to Project Emporia \cite{PE11} users in real-time. The news on user selected topics are coming from RSS feeds and tweets. The main challenge consists in training the underlying classifier on-line, as the news' trends have to be promptly classified. This problem is solved in two steps: First, appropriate unlabeled tweets are selected by means of AL and forwarded to Amazon Mechanical Turk (AMT) \cite{AMT} for labeling. Second, a Bayesian corroboration model, that takes the reliabilities of AMT workers into account, is used to label the corresponding tweets.
	
	Microsoft and Yahoo Lab present a further application of AL for detecting and filtering abusive user-generated content (UGC) on the Web in \cite{CZL+11}. The huge number of posts and the user's ability to learn how to avoid static, hand-coded detection rules, motivates the need of fast, automatic detection systems, that periodically update based on newly labeled data, provided by paid annotators. Therefore, AL is used to reduce the labeling costs.
	
	IBM uses AL in \cite{MS09} to carry out a sentiment analysis of texts, i.e., an automatic analysis of emotions expressed in posts. The goal is to identify whether a post reveals positive or negative opinions, e.g., criticism or approval.
	
	\item \textit{Speech Recognition:} In \cite{HRT06,TSH03} AT\&T shows how AL techniques can be employed to reduce the effort of transcribing natural spoken language required to generate labeled instances for data-driven speech and language processing systems. The goal of such a system may be to automate the classification of customer calls according to their content (e.g, account balance or rates inquiries) and consequently work through the customer requests with the help of a dialog. The transcription of an utterance is usually performed by a human expert and is very expensive, so the costs for extension and improvement can be reduced by an AL approach. At first, the system is trained based on an initial set of transcriptions and, in the following steps, it iteratively selects from a large pool of utterances those with a low transcription confidence. These utterances are then transcribed by an expert and the system is updated.
	
	Another approach to minimizing the transcription effort is presented by Microsoft Research in \cite{YDA10}, which combines AL with semi-supervised learning to select most likely misclassified as well as frequently occurring, untranscribed utterances.  
	
	\item \textit{Image Classification:} Mitsubishi Electric Research Laboratories present in \cite{JPP12} an AL approach to minimize the image classification cost. The novelty of the presented approach consists in its possibility to start the training process without being aware of the total number of classes (categories). Initially, a classifier is trained based on a small amount of labeled images. Then, a pair of an unlabeled and a labeled image is selected and presented to the human expert for a ``match'' or ``no-match'' response. In case of a ``match'', the queried image is labeled accordingly (the same category as the labeled image), added to the labeled set and the classifier is updated. Otherwise, based on the selection algorithm another labeled sample is selected for comparison. The query step is repeated until the ``match'' response is obtained or the unlabeled sample does not match any of the labeled samples. In the latter case, the number of classes is increased and a new category (label) is assigned to the queried sample.
	
	\item \textit{Drug Design:} Quantitative Medicine\footnote{\url{http://qtmed.com/home-page/active-learning/} -- last access 02/24/2015} claims to offer the first AL software as a service (SaaS) solution for drug discovery and development. By using AL techniques the efficiency of identifying optimal drug candidates is substantially improved.
	
	\item \textit{Malware Detection:} In order to detect and remove malware (malicious software) most anti-virus vendors rely on signature-based detection of previously seen malwares. Therefore, they have to keep their signature repository up to date by collecting suspicious files, which are then manually analyzed by security experts. As the labeling process is a time-consuming task, AL techniques may help the security experts to reduce the number of manually inspected files.
	Telekom Innovation Laboratories present in \cite{NMR+14} an AL framework to select the most likely to be malware files for manual inspection. Support vector machines (SVM) with radial basis function (RBF) kernel are used as classification algorithm. The most informative samples are acquired in two phases: First, the files for which the classifier is most unconfident (low distance to the separating hyperplane) about the category (malware or benign) are selected for manual analysis. In the second phase, the files that reside deep inside the malicious side (high distance to the separating hyperplane) are	picked out for further investigation. The proposed AL techniques was tested in a 10 day experiment, showing an improvement in the daily number of new detected malwares.

	\item \textit{Recommender Systems:} The Android application Shopr \cite{Trottmann15} is a mobile recommender system that suggests clothing items from stores situated in user's vicinity. The system requires feedback from users in order to understand and fulfill his/her desires. At this point comes AL into play narrowing down the item space to those that match the user's interest, taking both similarity and dissimilarity between the items into account. For a detailed description of the AL-based recommender system see \cite{LTW14}.			
			
\end{itemize}

	In the past years there is a strong increase of devices connected to internet which generate a storm of raw data, making the term \textit{Big~Data} become popular. From a data analytics point of view, big data may be defined by the four~\textit{V}'s: \textit{V}olume, \textit{V}ariety, \textit{V}elocity and \textit{V}eracity. Volume describes the big size of data, Velocity indicates fast generating rates, Variety implies the heterogeneity of data, and Veracity refers to the uncertainty of data quality \cite{ZCJ+14, BigData15}. 
	Further, the \textit{Internet of Things} aims at connecting everything with everything, leading to a huge network of things that should possess a certain intelligence that may support and improve the daily life of people around the world. To extract high-quality information from big data it may necessary for the system to receive feedback, e.g, from domain experts. Moreover, the humans should benefit too, by exchanging and improving their knowledge. 
	With regard to the previously presented trends, CAL may contribute to addressing these new, future challenges.

Altogether, we can be sure that there will also be an increasing interest in AL and, as many limitations of AL are abolished, in CAL, too.

In the remainder of this article, we first present some foundations of AL in Section \ref{sec:Foundations} and summarize results of own, preliminary work in Section \ref{sec:PreliminaryWork}. Section \ref{sec:casestudy} presents some experimental results for our AL techniques. In Section~\ref{sec:Challenges} we investigate the above challenges of CAL in more detail and briefly discuss possible solutions. Finally, Section \ref{sec:SummaryOutlook} concludes the article by taking a look at possible application fields.

\section{Overview of Active Learning Foundations}\label{sec:Foundations}

The motivation of AL is that obtaining plenty of unlabeled data is often quite cheap, while acquiring labels is a task with high costs (monetary or temporal). AL is based on the hypothesis that a process of (iteratively) asking an \textit{oracle} for labels and refining the current model can be realized in a way such that
\begin{itemize}
	\item the performance of the resulting model is comparable to the performance of a model trained on a fully labeled data set and
	\item the overall labeling costs to obtain the final model are much lower (typically simply measured by the number of labels).
\end{itemize}
Actually, to address the previous requirements it is possible to build an \textit{active learner} that is based on a complementary pair of \textit{model} (e.g., a classifier) and \textit{selection strategy}. With a selection strategy, the active learner decides whether a sample is \textit{informative} and asks the oracle for labels. Here, informative means that the active learner expects a (high) performance gain if this sample is labeled (similarly, a set of samples can also be called informative). 

\todo{BS@BS: Neues Paper aus Google Scholar zur Klassifikation von Blattkrankheiten einbauen.}
Basically, various kinds of models can be used for AL, but the selection strategy should always be defined depending on the model type (e.g., whether SVMs, neural networks, probabilistic classifiers, or decision trees are chosen to solve a classification problem). In the remainder of this article we focus on AL for classification problems (see \cite{RS13,RCS14,CL06,ZYPPJC15}, for example), but AL can be applied to modify the results of clustering (see \cite{MCR12}, for example) or to regression problems, too (see \cite{CZZ13,PM11,DB14,DMB13}, for example).

\begin{figure}[htb]
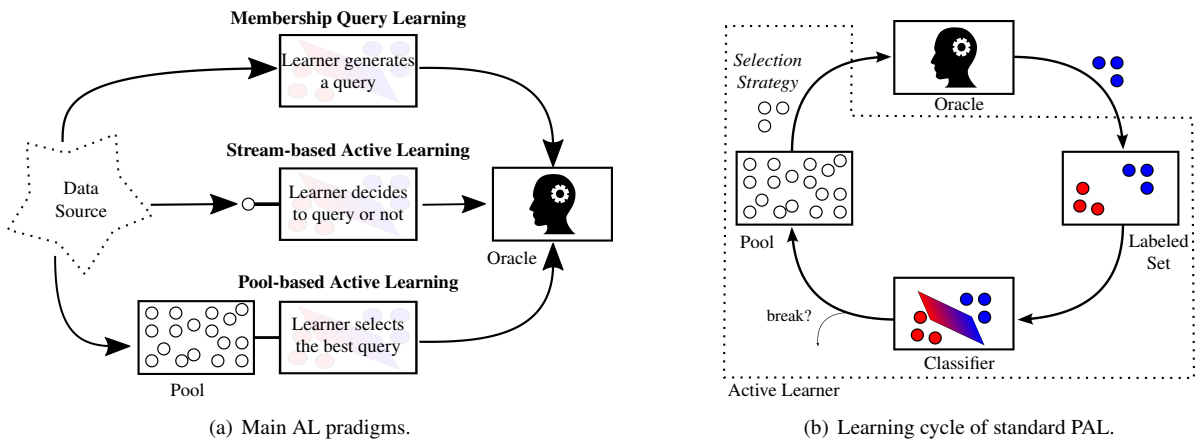

	\centering
	\subfigure[Main AL pradigms.]{\scalebox{.7}{\input{images/al_scenarios.tex}}}
	\hspace*{10mm}
	\subfigure[Learning cycle of standard PAL.]{\scalebox{.7}{\input{images/pool_based_cycle.tex}}}
	\caption{Overview of main AL scenarios with focus on PAL.}
	\label{fig:PAL}
\end{figure}

In the field of AL, membership query learning (MQL) \cite{Angluin88}, stream-based active learning (SAL) \cite{ACLEM90}, and pool-based active learning (PAL) \cite{LG94} are the most important paradigms (see Figure \ref{fig:PAL}, left hand side).
 
In an (MQL) scenario, the active learner may query labels for any sample in the input space, including samples generated by the active learner itself. Lang and Baum \cite{LB92}, for example, describe an (MQL) scenario with human oracles for classifying written digits. The queries generated by the active learner turned out to be some mixtures of digits, therefore being too difficult for a person to provide reliable answers.

An alternative to (MQL) is (SAL), which assumes that obtaining unlabeled samples generates low or no costs. Therefore, a sample is drawn from the data source and the active learner decides whether or not to request label information. Practically, in an (SAL) setting the source data is scanned sequentially and a decision is made for each sample individually. The SAL scenario has been applied, for example, in \cite{Yu2005} to learn ranking functions for retrieving for-sale houses listed on realtor.com that meet the user's preferences.

Typically, SAL selects only one sample in each learning cycle. 

For many practical problems a large set of unlabeled samples may be gathered inexpensively and this set is available at the very beginning of the AL process. This motivates the PAL scenario. The learning cycle of PAL is depicted in Figure~\ref{fig:PAL} on the right hand side. Typically, PAL starts with a large pool of unlabeled and a small set of labeled samples. On the basis of the labeled samples the active learner is trained. Then, based on a selection strategy, which considers  the ``knowledge'' of the learner, a query set of unlabeled samples is determined and presented to the oracle (e.g., a human domain expert), who provides the label information. The set of labeled samples is updated with the newly labeled samples and the learner updates its knowledge. The learning cycle is repeated until a given stopping condition is met. 

In the remainder of this article we focus on PAL for the sake of simplicity. Many ideas, however, may be transferred to MQL or SAL.

A selection strategy for PAL has to fulfill several tasks, two of which shall be given as an example: At an early stage of the AL process, samples have to be chosen in all regions of the input space covered by data (\textit{exploration phase}). At a late stage of the AL process, a fine-tuning of the decision boundary of the classifier has to be realized by choosing samples close to the (current) decision boundary (\textit{exploitation phase}). Thus, ``asking he right question'' (i.e., choosing samples for a query) is a multi-faceted problem and various selection strategies have been proposed and investigated. We want to emphasize that a successful selection strategy has to consider structure in the (un-)labeled data. We come back to this point later.

Typically, the very limiting assumptions listed in Section \ref{sec:Introduction} are made concerning the oracle and the labeling costs (omniscient, omnipresent oracle that labels samples on a fixed cost basis). Moreover, some other aspects of real-world problems are often more or less neglected by current research, for example:
\begin{itemize}
	\item In real-world applications, AL has often to start ``from scratch'', i.e., with no labels at all. This requires sophisticated selection strategies with different behaviors at different phases of the AL process.
	\item Parameters of the active learner (including parameter of training algorithms of the classifier and the selection strategy) cannot be found by trial-and-error. AL only allows for ``one shot''.
\end{itemize}

There are several articles that assess the state of the art in AL. We do not want to replicate this work here but refer to the (in our opinion) most important articles:
\begin{itemize}
	\item A general introduction to AL, including a discussion of AL scenarios and an overview of query strategies is provided in \cite{Set09}.
	\item A detailed overview of relevant PAL techniques is part of \cite{RS13}. In addition to single-view/single-learner methods, alternative approaches are outlined: multi-view/single-learner, single-view/multi-learner, and multi-view/multi-learner.
	\item For certain problem areas it makes sense to use AL in combination with semi-supervised learning (SSL). AL techniques that integrate SSL techniques are succinctly presented in \cite{RCS14}.
	\item AL in combinations with SVM for solving classification problems are summarized in \cite{JH09, KPI14}.
\end{itemize}


\section{Preliminary Work -- \textit{What are we able to do right now}?}\label{sec:PreliminaryWork}

In our preliminary work we tried to overcome some of the mentioned limitations. In particular, we focused on
\begin{enumerate}
	\item capturing structure in (un-)labeled data to support the exploration/exploitation phases of new selection strategies for PAL,
	\item self-adapting weighting schemes for different criteria combined in a selection strategy, and
	\item parameter-free AL (apart from parameters that can be found offline using the unlabeled data before the AL process starts, for instance).
\end{enumerate}
We still make the (also very limiting) assumptions concerning oracles and labeling costs that will be addressed later (cf.\ Section \ref{sec:Challenges}). 
Solutions to the above challenges were not developed separately but with a holistic view on the problem field. Thus, the following subsections sketch closely related solutions and refer to the publications where these approaches are described in much more detail.


\subsection{Capturing Structure in Data}

%

Machine learning becomes possible if certain ``regularities'' or ``patterns'' in data can be identified and exploited. For classification problems we may assume that the data form clusters of arbitrary shape. For AL, such structure in data has to be captured to improve selection strategies and/or to improve the training of the classifier. In our work, we made two important assumptions that will be outlined in the following. We want to emphasize first that we do not claim that these assumptions always hold. They hold often and to a certain degree and, thus, building solutions based on these assumptions leads to an average, but yet significant improvement of a classifier's performance (shown on average over a number of benchmark data sets, see Section \ref{sec:casestudy}).

First, we assume that there is a correspondence of processes in the real world that ``generate'' the samples that we 
observe and want to classify on the one hand, and clusters in the training data on the other hand. We may assume further that processes are uniquely assigned to classes and, thus, clusters can also be uniquely assigned to classes. Of course, this does not contradict the fact that clusters belonging to different classes might widely overlap, which means that they cannot be clearly identified if we have the unlabeled data only.

Second, in the real world, these data-generating processes (and the mechanisms necessary to observe these processes) are affected by a (superposition of) random influences (or influences that we have to regard as being random). Two examples are (1) stochasticity which is inherent to certain processes or (2) measurement effects such as sensor noise. Following the generalized central limit theorem \cite{DHS01} we may assume that structure in continuous (real-valued) input dimensions of a classifier may well be captured by means of (mixtures of) Gaussian (i.e., normal) distributions, called components of this mixture model. Apart from that reason, under some mild assumptions it is always possible to model continuous densities using mixtures of Gaussians with arbitrary precision. In practical applications, particular discrete dimensions (integer dimensions) can often be handled like real ones.

Based on these assumptions we developed solutions in an evolutionary approach.

\begin{enumerate}
\item \textit{Capture Data Structure in AL:} In a first step, we decided to use probabilistic generative models to capture structure in data and to build classifiers based upon these models. The probabilistic models are parametric density models, mixtures of Gaussians for continuous (or integer where appropriate) input dimensions and special cases of multinomial distributions for the categorical ones. These models can be parametrized (trained) from a set of unlabeled data either in an expectation maximization approach or in a Bayesian approach, called variational Bayesian inference (VI) \cite{Bis06}. The VI has the advantages that, in contrast to EM, effects caused by singularities can be avoided and the number of components in the mixture models can be determined automatically with a built-in pruning technique \cite{FS09}.
Having found the density model, a classifier (CMM: classifier based on mixture models) is constructed using any available labels. This classifier gradually (i.e., with a certain probability) assigns the model components to classes. The components are intended to model the data originating from the data-generating processes in the real world. Our approach for AL based on CMM now exploits the density information in various ways (see below). It was first proposed in \cite{RS11} and extended in \cite{RS13}.

\item \textit{Revise Captured Structure during AL:} The approach above has the property that the density models are found in an unsupervised way using the unlabeled data available at the very beginning of the AL process. Any label information becoming available throughout the AL process is not used so far to refine the model. With label information, for example, overlapping clusters assigned to different classes could be identified more easily. The approach above is called $\text{CMM}_{sha}$ (shared-components classifier) because in an unsupervised training approach all classes ``share'' the same density model. In a supervised approach, separate density models are trained for the different classes and then combined. This leads to the $\text{CMM}_{sep}$ (separate-components classifier) which may basically perform better in many applications regarding data modeling and classification accuracies. In an AL approach, we must start with a $\text{CMM}_{sha}$, but, when more and more labeled data become available, we may iteratively transform this $\text{CMM}_{sha}$ towards a $\text{CMM}_{sep}$. In a second step, we realized this idea by adopting techniques from non-parametric density estimation, nearest neighbor classification \cite{RC14}, and transductive learning. The latter is also a variant of semi-supervised learning where labels are found for unlabeled data using the labeled fraction of data. 
More details about this AL technique can be found in \cite{RCS14}. 
Preliminary work has shown that the samples queried by means of this techniques are not completely biased to the actively trained CMM$_{sep}$ and can be (re-)used to train a different classifier paradigm as well, e.g., classifiers such as support vector machines (SVM).


\item \textit{Exploit Structure for AL with SVM:} In principal, generative classifiers such as CMM often perform worse than discriminative classifiers such as SVM in many applications. But, on the other hand no density information (or, more general, information concerning structure of data in the input space of the classifier) can be extracted from standard SVM to use it in selection strategies (see below). Thus, a first idea would be to build a generative data model and a discriminative classifier in parallel in an AL process. Having investigated this idea first, we then decided to follow another idea in a third development step of our AL technique: 
We developed the responsibility weighted Mahalanobis (RWM) kernel \cite{RS15}, a new kernel for SVM that assesses the similarity of samples by means of a Mahalanobis distance in the case of Gaussian mixtures. Thereby, model components that are assumed to be ``responsible'' for the generation of a sample get a high weight. The RWM kernel can be used for AL with SVM in combination with various selection strategies. This third evolution step in our approach to capture structure in data is part of our ongoing research.
\end{enumerate}


\subsection{Self-Adapting Selection Strategies}

A key component of an AL process is the selection strategy. Uncertainty sampling (US) strategies, for example, are frequently used in AL processes. The idea is to select the sample for which the classifier (or a committee of classifiers) is most uncertain concerning its class assignment. This approach has some drawbacks: As the queried samples are always close to the (current) decision boundary, the exploration of the input space may be suppressed, and if more than one sample is queried in each learning cycle, then the selected samples are similar to each other. A selection strategy should be able to detect all decision regions (exploration phase) and fine-tune the decision boundary (exploitation phase), which means that an efficient and effective selection strategy has to find a good trade-off between exploration and exploitation. Thus, our approaches for selection strategies are based on the following two hypotheses:
 \begin{enumerate}
	 \item A selection strategy has to consider various aspects and, thus, must combine several criteria.
	 \item In different phases of the AL process, these criteria must be weighed differently.
 \end{enumerate}

In a first step, we developed the selection strategy 3DS which combines three criteria \cite{RS11}:
\begin{enumerate}
	\item the \textit{density} of regions where samples are selected, 
	\item the \textit{distance} of samples to the decision boundary, and
	\item the \textit{diversity} of samples in the query set.
\end{enumerate}
The density is an exploration criterion, the distance is an exploitation criterion, and the diversity has to be considered for query sizes larger than one to avoid asking for redundant information (i.e., for efficiency reasons). These criteria can be weighed individually in a linear combination. Moreover, we developed a self-adaptation scheme for 3DS that 
weights the density criterion more strongly at the beginning of the AL process, in order to explore different regions. In later cycles it is necessary to exploit the gathered information, therefore the distance criterion is emphasized.  

In a second step, we extended the 3DS strategy by another criterion:
\begin{itemize}
	\item the class \textit{distribution} of samples is considered indirectly by evaluating responsibilities.
\end{itemize}
That is, this 4DS strategy (see \cite{RS13} for details) aims at labeling samples in a way such that the distribution of the samples approximates the unknown true class distribution. 
This is especially beneficial for data sets with an unbalanced class distribution, as the generalization performance is improved. How can this be done as we do not know the labels in advance? This is possible (1) by assuming that processes in the ``true'' world can uniquely be assigned to classes (see above) and (2) by considering the responsibilities of model components (that model these processes) for the samples. Responsibilities are estimates of conditional probabilities, that indicate how likely specific processes modeled by corresponding components are ``responsible'' for ``the generation of a given sample'' (i.e., that the sample originates from the considered processes). 

The self-adaptation of weights in 3DS was extended in 4DS with the idea of focussing on the class distribution criterion in initial cycles of an AL process.


\subsection{Parameter-free AL}

In an AL process, many parameters have to be set: parameters of learning algorithms for classifiers, parameters of selection strategies, etc. Typically, a real application of AL only allows for ``one shot'' for some of these parameters. The selection strategy, for example , should be parameter-free. Other parameters can be tuned, e.g., those of techniques that capture structure in unlabeled data before the AL process starts. 

In our work we addressed the following parameter types:
\begin{itemize}
	\item \textit{Parameters of techniques needed to capture structure in unlabeled data:} Appropriate parameters of the VI algorithm (see above) can in principal be found by repeated training and analysis of reached likelihood values, for instance. Another possibility is to analyze the representativity measure, as described in \cite{FKSO15}. 
	\item \textit{Parameters of the selection strategies:} Here, we realized the idea that the active learner should be free of such parameters. In 4DS (and 3DS) we start with appropriate initial weights of criteria in the linear combination and let the system self-adapt these weights (see above). A parameter that still remains as it has definitely to be set by a user is the query size. For a query size of one sample, 4DS already is parameter-free. For larger queries we still have to set the weight for the diversity criterion. Finding appropriate heuristics to set this parameter is part of our current research.
	\item \textit{Parameters of algorithms for classifier training:} Having found the density model, e.g., by means of VI, no further parametrization is required for constructing the CMM classifier, which can be trained using any available label information.
	The parametrization of SVM is part of our ongoing research. Here, we also aim at adapting the penalty factor $C$ (of $C$-SVM) and the kernel width $\gamma$ (of Gaussian kernels) considering the observations made while applying the parametrization heuristic for $C$ and $\gamma$ presented in \cite{KL03}.
\end{itemize}

\subsection{Summary of Preliminary Research}

We could show that our AL approach is able to boost the classification accuracy significantly. Here ``significantly'' actually means that we applied statistical tests to show the superiority of our techniques on certain significance levels. Apart from accuracy measures (to assess the effectiveness of our AL approach) such as the \textit{ranked performance} on a number of benchmark data sets, we also applied other measures to assess the efficiency of our AL approach (i.e., the learning speed) such as a \textit{data utilization rate} or the \textit{area under the learning curve} \cite{CKS06}. 
We also defined a new measure, the \textit{class distribution match} \cite{RS13}. 

Some (in our opinion) less important assumptions or achievements were not mentioned so far. Examples are the assumption that the computational costs of AL are negligible compared to labeling costs or the fact that our AL approach is in principal able to start the AL process with a completely unlabeled data set (in contrast to many other approaches).


\section{Preliminary Experimental Results}\label{sec:casestudy}

In this section we will compare our AL technique with RWM kernel and 4DS selection strategy to an AL technique with generative classifiers, $\text{CMM}_{sha}$ with 4DS strategy, and to an AL technique with discriminative classifiers, SVM with RBF (Gaussian) kernel and uncertainty sampling (US) strategy. First, we visualize the behavior of the two kernels on a two dimensional data set. Second, simulation experiments are conducted and evaluated for 20 benchmark data sets.

\subsection{Behavior of SVM with RBF and RWM Kernels}

We decided to visualize the behavior of SVM with RBF and RWM kernels on the Clouds data set from the UCI Machine Learning Repository \cite{AN07}. We performed a $z$-score normalization, conducted a stratified 5-fold cross-validation and choose the first fold for presentation here. Figures \ref{fig:USRBF} and \ref{fig:RWM4DS} show the state of SVM with RBF and RWM kernels at different cycles of the AL processes. The orange colored samples correspond to the 8 initially selected samples, whereas the actively selected samples are colored red if they are selected in the current query round otherwise purple. The support vectors are indicated by framing the specific samples with a black square. The decision boundary is depicted as a solid black line. In case of the RWM kernel, Figure \ref{fig:RWM4DS}, the gray colored ellipses correspond to level curves of the Gaussians the RWM kernel is based on (located at centers indicated by large $\times$s). Points on these level curves have a Mahalanobis distance of one to the center of the respective Gaussian.

We can see that the SVM with RBF kernel performs worse at the beginning of the AL process, as it does not exploit the unlabeled data. By using structure information derived from the unlabeled data, the RWM kernel achieves a noticeably higher classification accuracy, which is maintained until the end of the AL process. 

Furthermore, we can state that US selects samples near the decision boundary (e.g., Figure \ref{fig:RBFCycle112}), whereas the 4DS selection strategy by its explorative manner selects samples in other regions, too (e.g., Figure \ref{fig:RWM4DSCycle112}).

\newcommand{\prefWidth}{0.24}
\begin{figure}[H]
	\begin{center}
		\subfigure[Cycle  0]{\label{fig:RBFCycle0}\includegraphics[width=\prefWidth\textwidth]{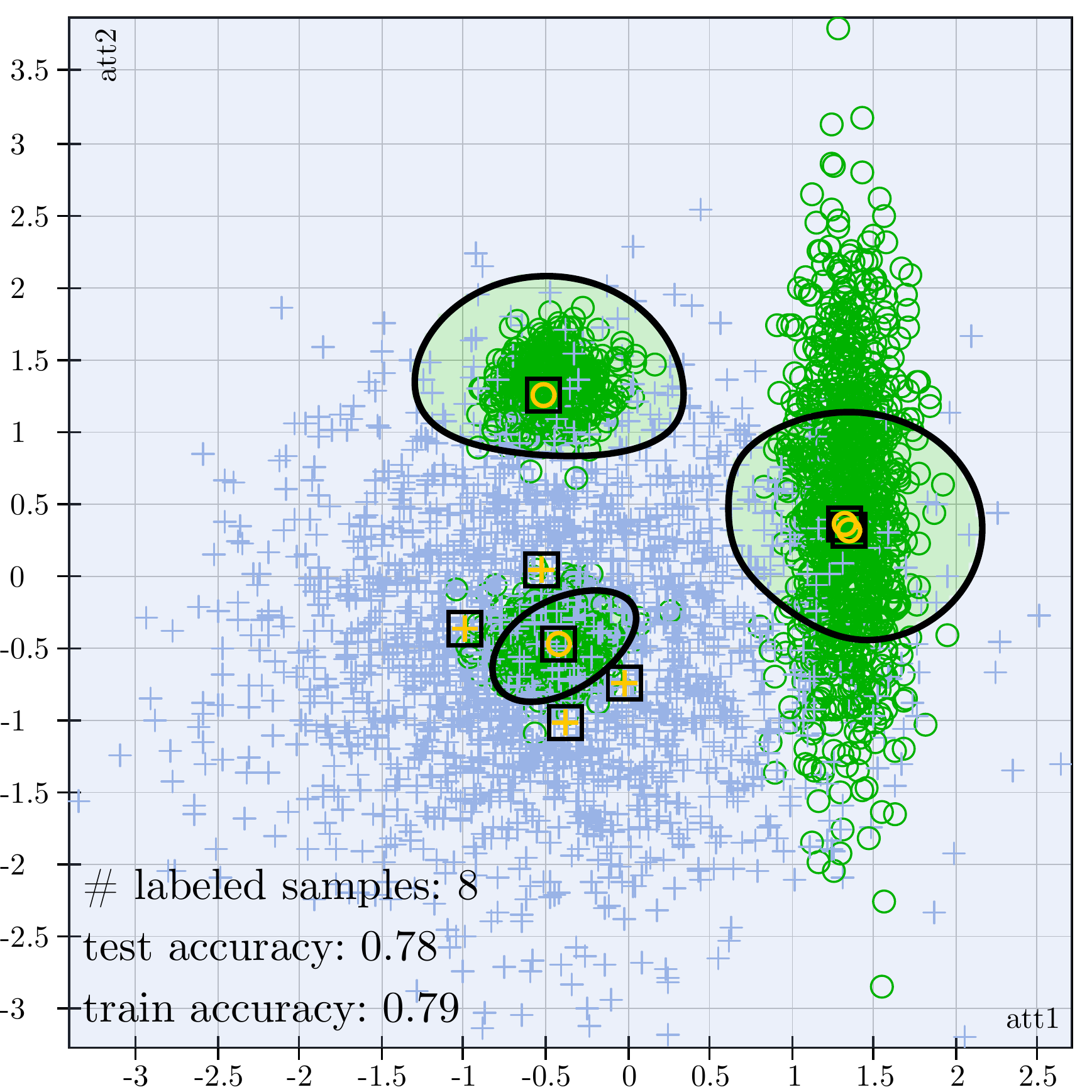}}
		\hfill
		\subfigure[Cycle  12]{\label{fig:RBFCycle12}\includegraphics[width=\prefWidth\textwidth]{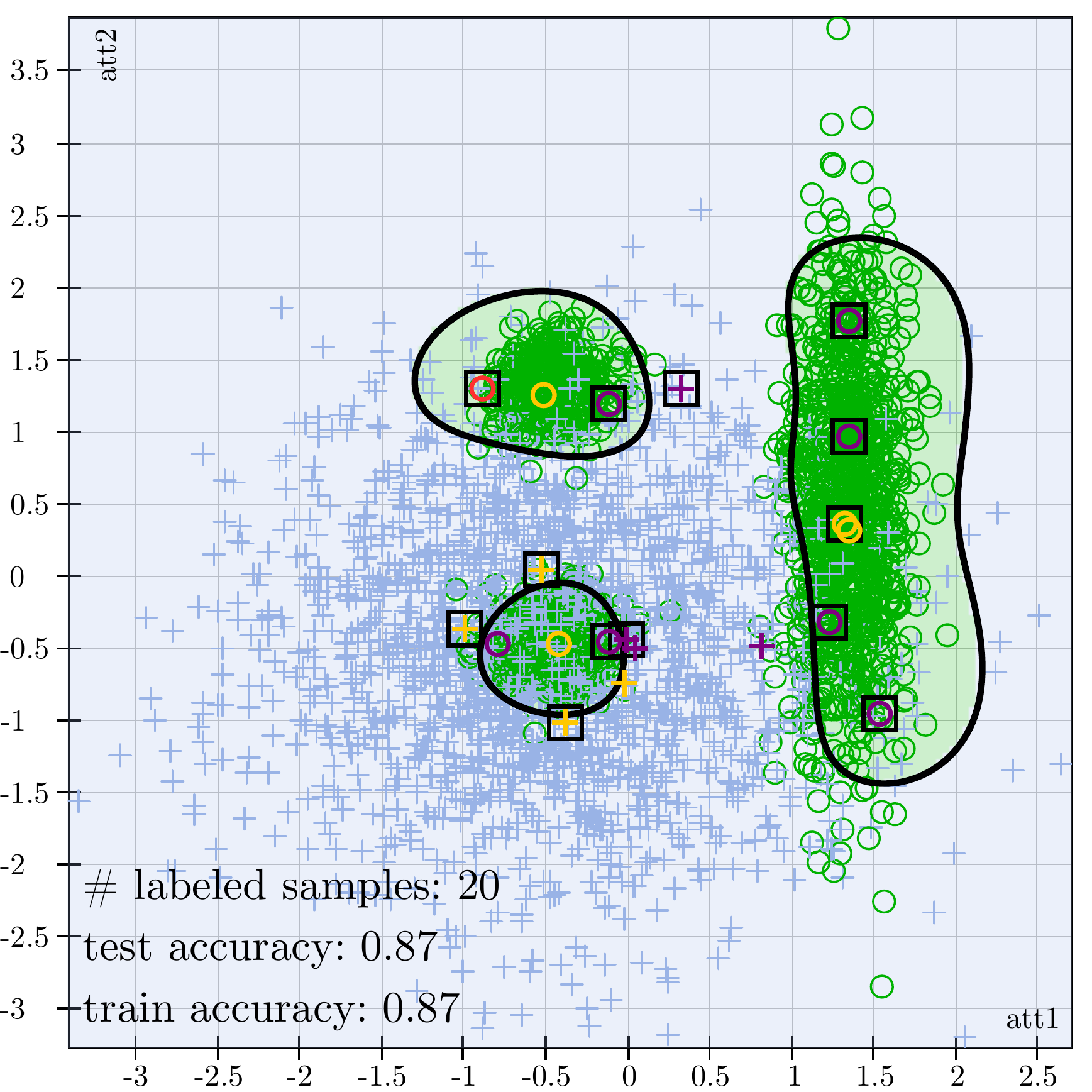}}
		\hfill
		\subfigure[Cycle  112]{\label{fig:RBFCycle112}\includegraphics[width=\prefWidth\textwidth]{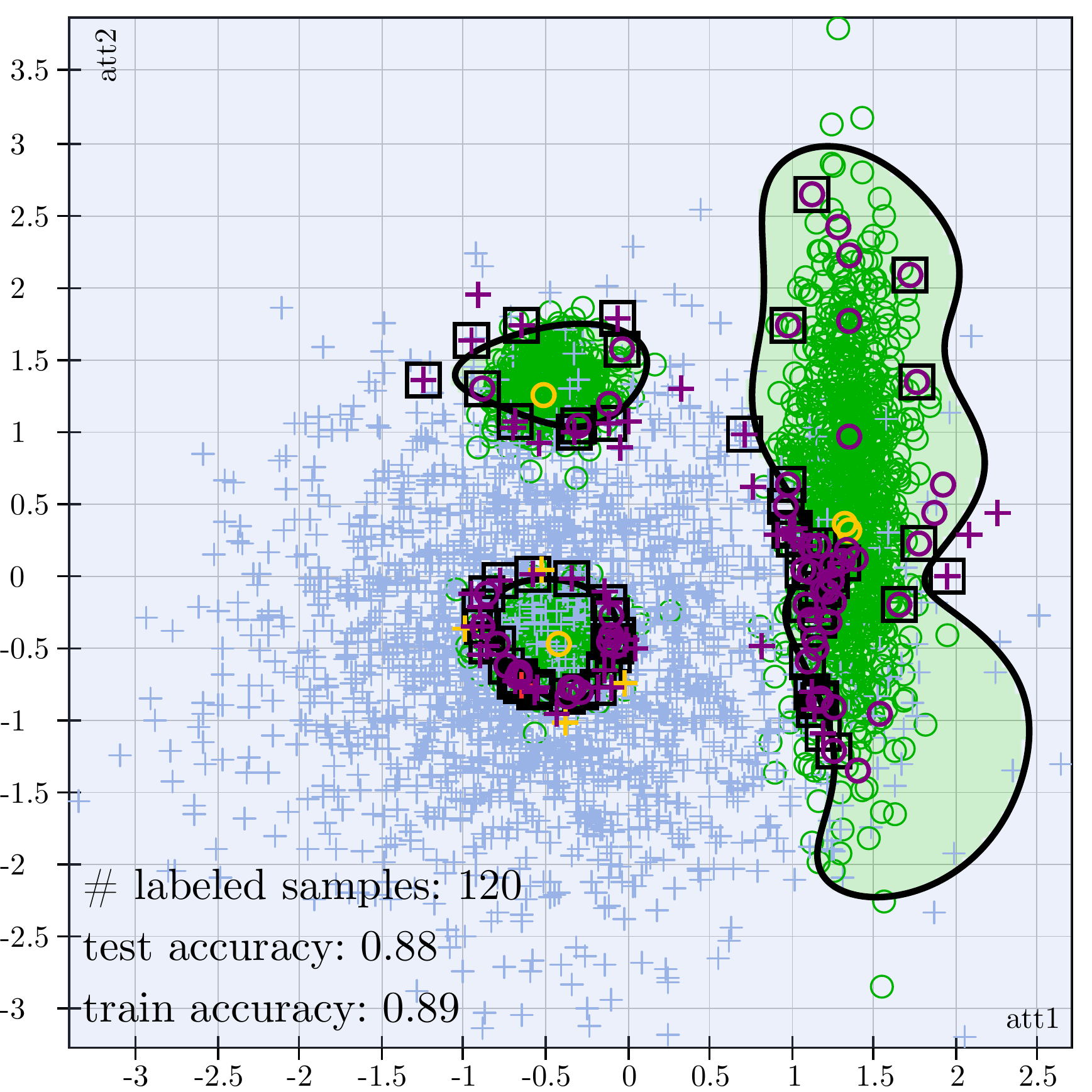}}
		\hfill
		\subfigure[Learning Curve]{\label{fig:RBFLearningCurve}\includegraphics[width=\prefWidth\textwidth]{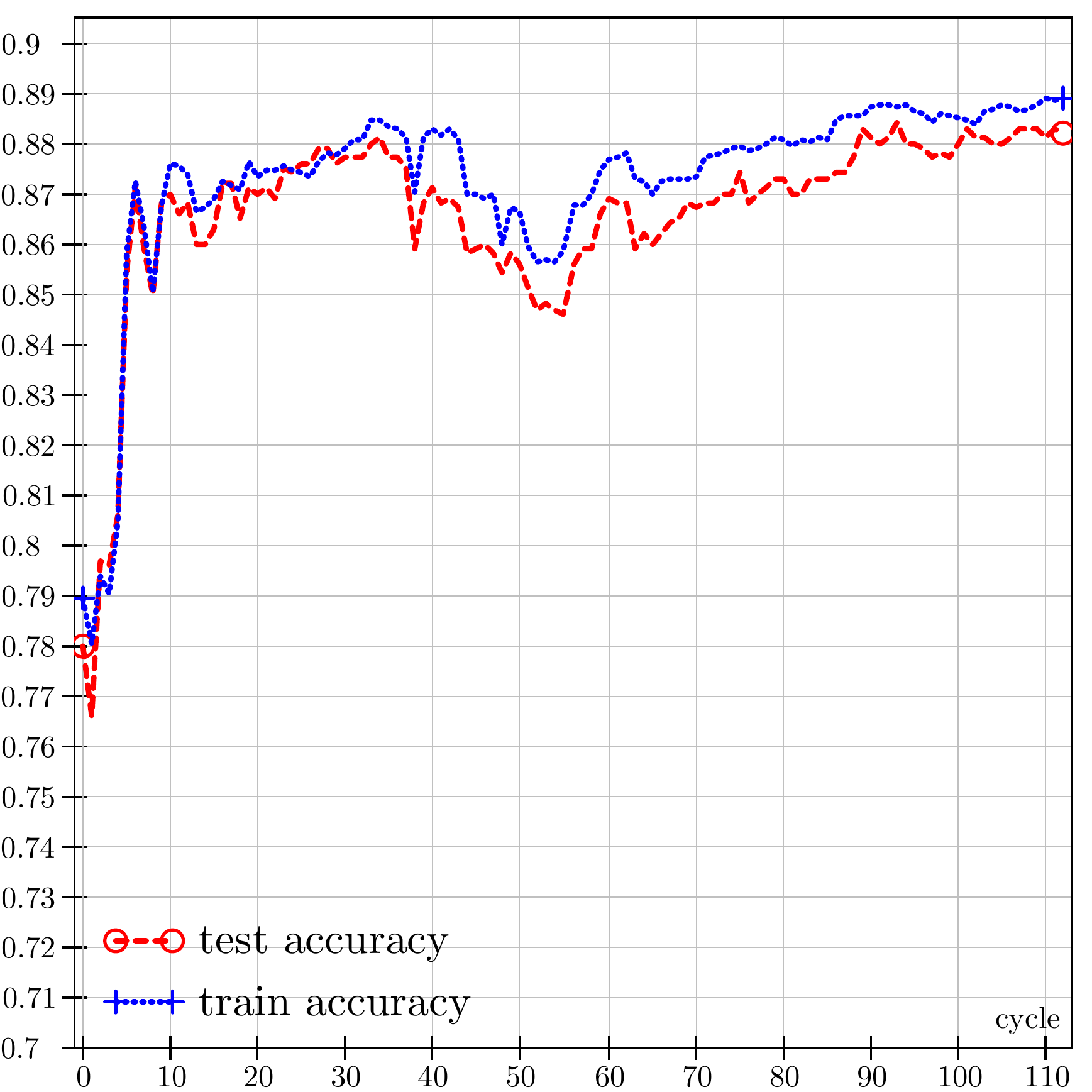}}
	\end{center}
	\vspace{-2.0em}
	\caption{Different stages of an AL process with SVM using RBF kernel and US selection strategy. The samples with known label are colored orange (initially selected) and red or purple (actively selected). The decision boundary is depicted as a solid black line.}
	\label{fig:USRBF}
\end{figure}

\begin{figure}[H]
	\begin{center}
		\subfigure[Cycle 0]{\label{fig:RWM4DSCycle0}\includegraphics[width=\prefWidth\textwidth]{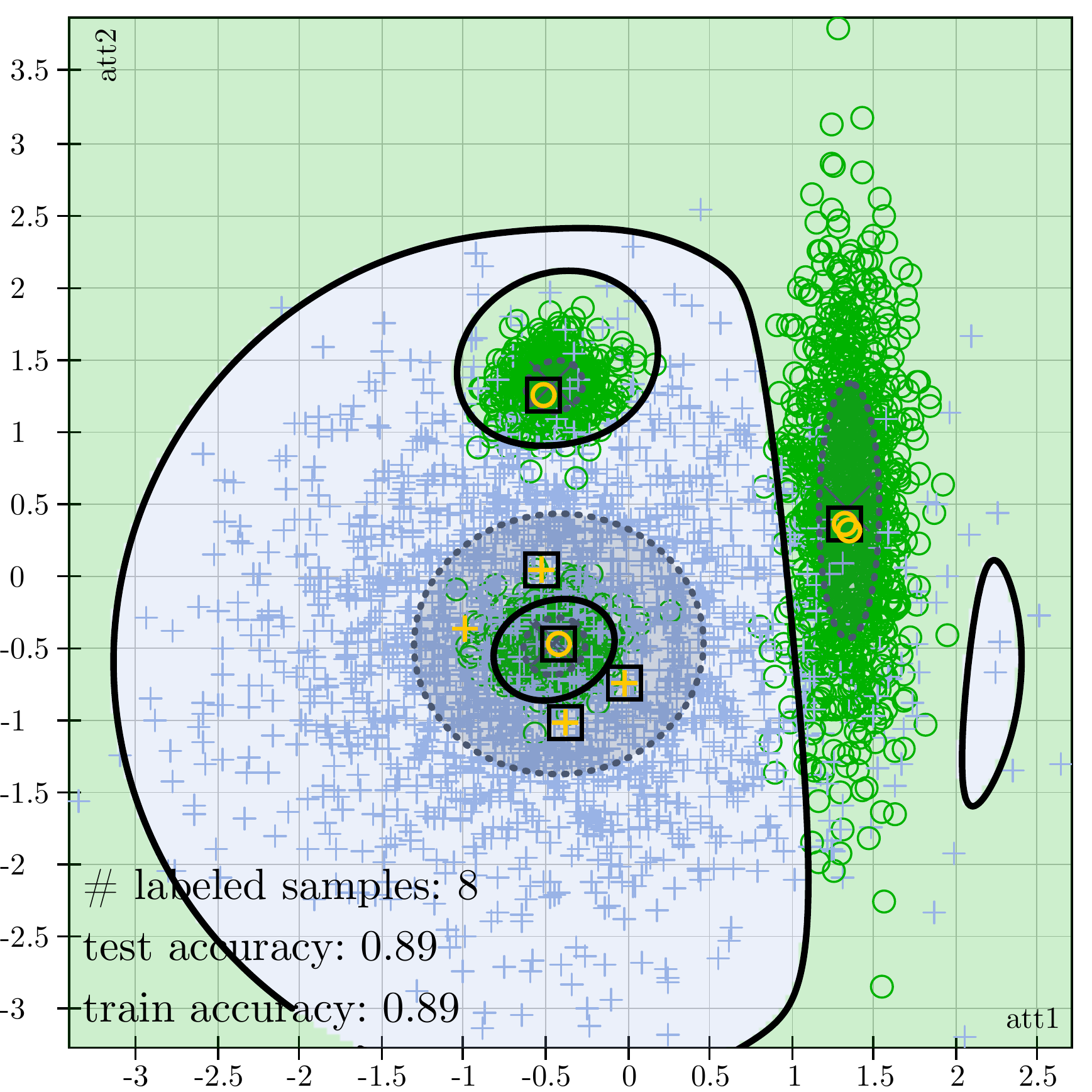}}
		\hfill
		\subfigure[Cycle 12]{\label{fig:RWMvCycle12}\includegraphics[width=\prefWidth\textwidth]{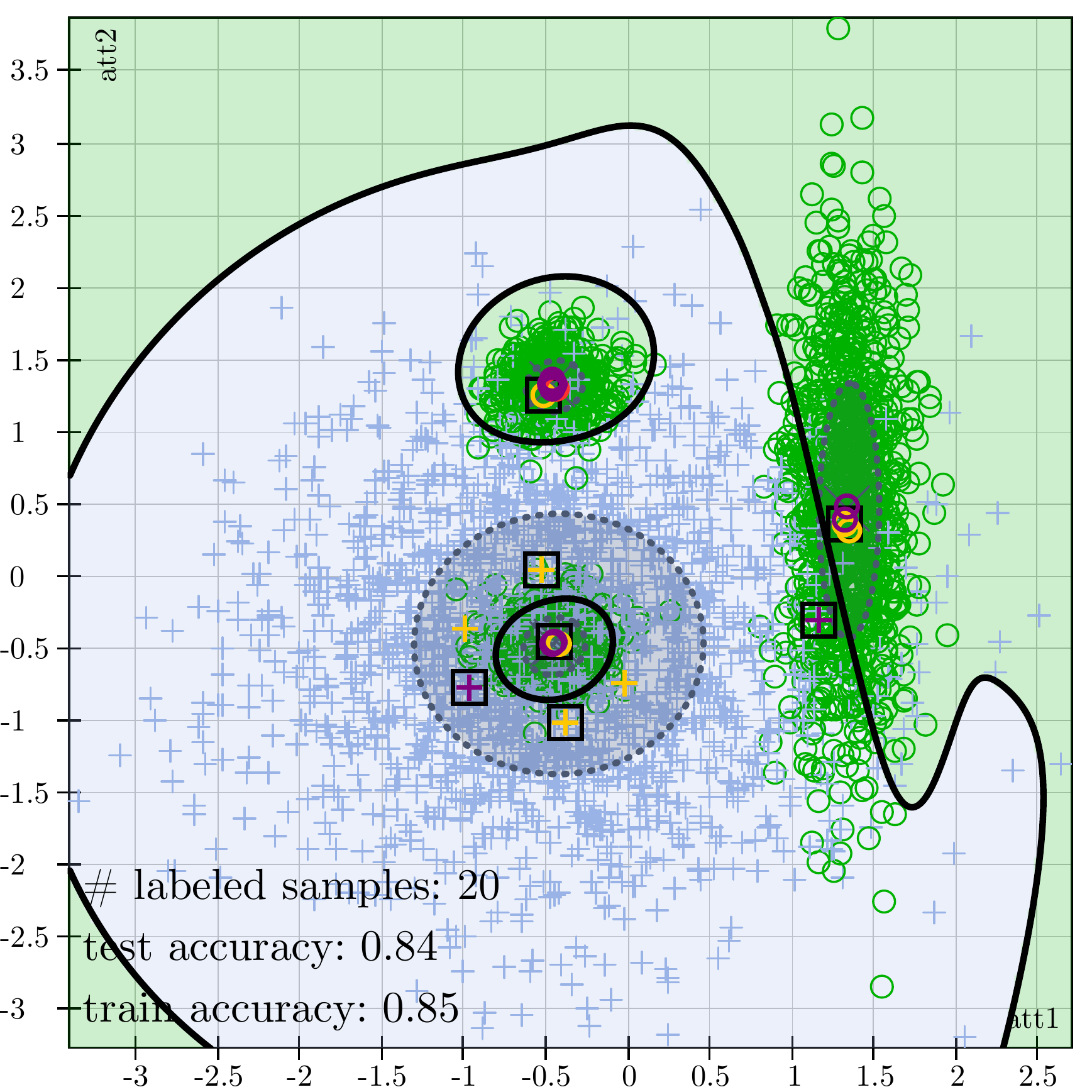}}
		\hfill
		\subfigure[Cycle 112]{\label{fig:RWM4DSCycle112}\includegraphics[width=\prefWidth\textwidth]{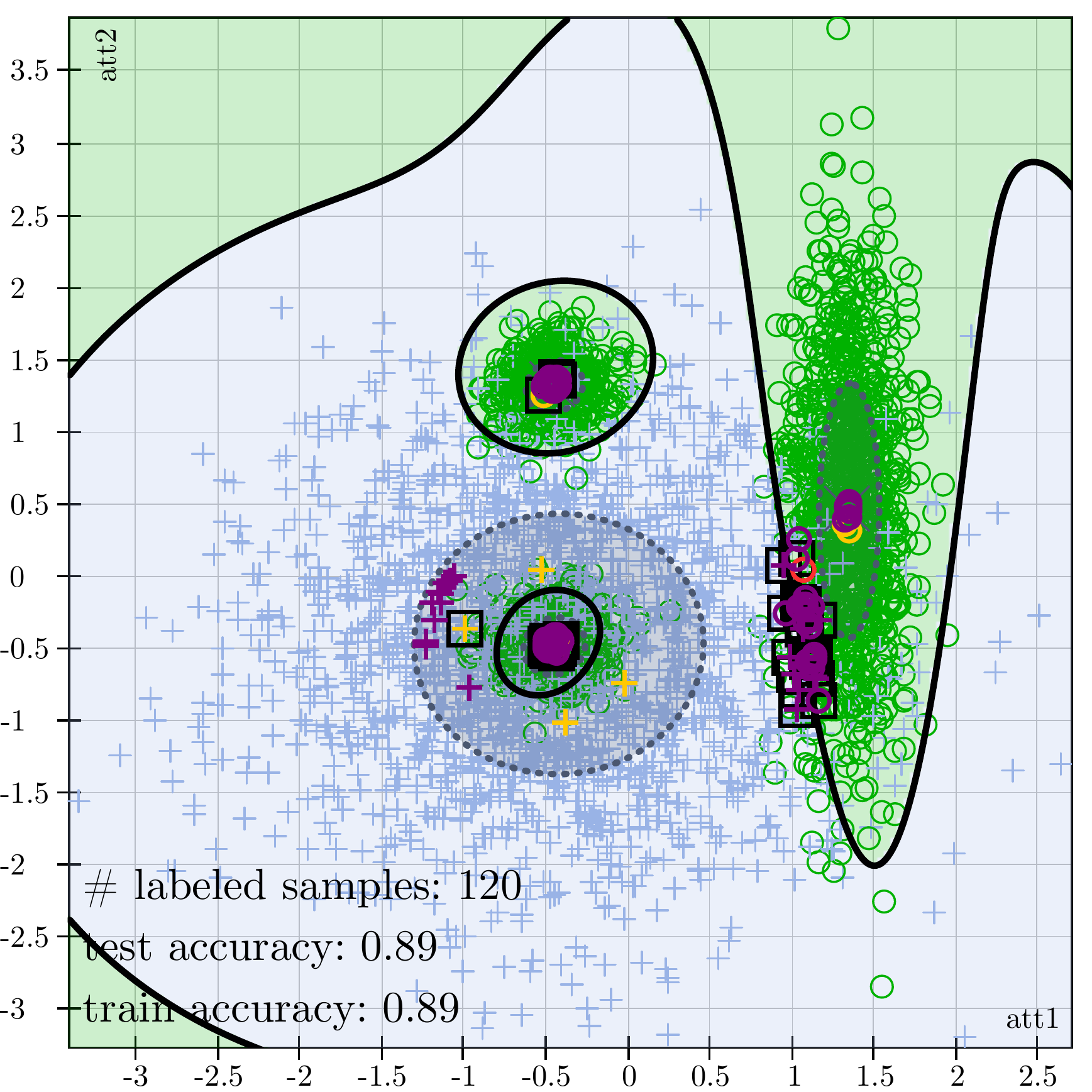}}
		\hfill
		\subfigure[Learning Curve]{\label{fig:RWM4DSLearningCurve}\includegraphics[width=\prefWidth\textwidth]{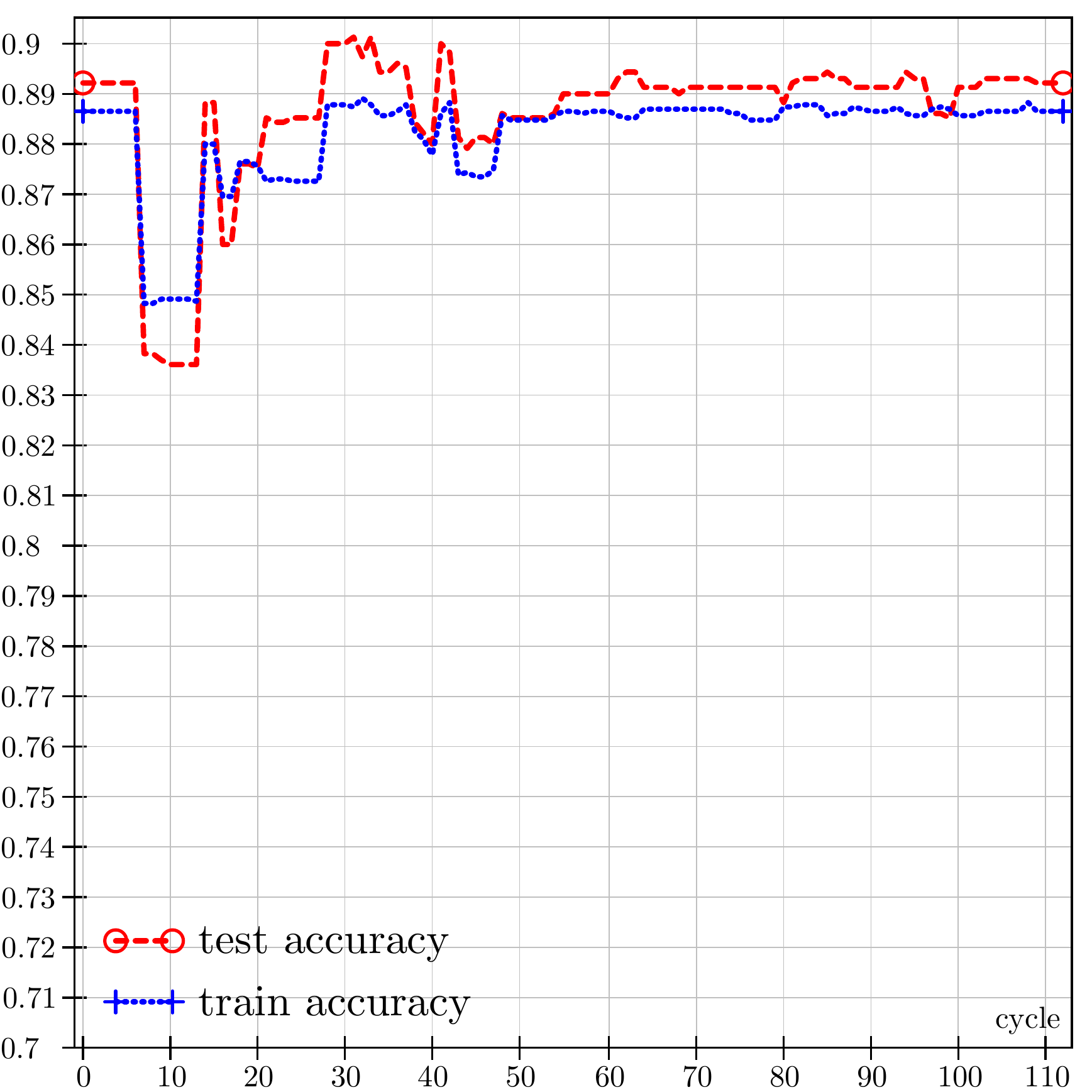}}
	\end{center}
	\vspace{-2.0em}
	\caption{Different stages of an AL process with SVM using RWM kernel and 4DS selection strategy. The samples with known label are colored orange (initially selected) and red or purple (actively selected). The decision boundary is depicted as a solid black line.}
	\label{fig:RWM4DS}
\end{figure}

\subsection{Results for 20 Benchmark Data Sets}

To evaluate the performance of the active learner with RWM kernel and 4DS selection strategy numerically we conduct experiments on 20 publicly available data sets. For more information regarding general data set characteristics and experimental setup see \cite{RCS14}. The AL techniques are ranked based on a Friedman test \cite{Fri40} with a significance value $\alpha$ of $0.01$ followed by a Nemenyi test \cite{Nem63} as post hoc test. A detailed description of the evaluation method can be found in \cite{RS15}. 

\begin{figure}[htb]
	\centering

		\begin{tikzpicture}[xscale=1]
		
		 \tikzstyle{every node}=[font=\scriptsize] 
		
		\node[anchor=south] at (1 + 0.490, 0.8) {CD = 0.980};
		\draw [|-|] (1, 0.8) -- (1 + 0.980, 0.8);;
		\draw (1, 0) -- coordinate (x axis mid) (3, 0);
		\foreach \x in {1, ..., 3}  \draw (4-\x, 0pt) -- (4-\x, 6pt) node[anchor=south] {\x};
		\foreach \x in {1, 1.5, ..., 3}
		\draw (\x, 0pt) -- (\x, 3pt);
		\matrix [matrix of nodes,row sep=0.01cm,column sep=2.5cm,column 1/.style={anchor=east},column 2/.style={anchor=west}] at (2.0, -.6)
		{
			\node (4DS CMM) {$\text{CMM}_{sha}$ with 4DS};  &  \node (4DS RWM) {RWM kernel with 4DS};\\
			\node (US MRBF) {RBF kernel with US};  & \\
		};
		\draw (4DS RWM)[anchor=east] -| (4-1.200, -0.05);
		\draw (US MRBF)[anchor=east] -| (4-2.200, -0.05);
		\draw (4DS CMM)[anchor=east] -| (4-2.600, -0.05);
		\draw[black, very thick] (4-2.600,-0.2) -- (4-2.200,-0.2);
		\end{tikzpicture}

	\vspace{-0.3em}
	\caption{Friedman test with significance value $\alpha$ of 0.01 followed by a Nemenyi post hoc test. Active learners that are connected do not achieve classification accuracies that are significantly different.}
	\label{fig:CDplot}
\end{figure}
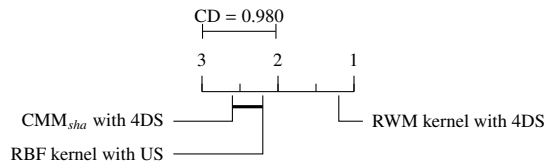

The classification accuracies achieved by each of the AL paradigms are shown in Table \ref{tab:Accuracies}. The average ranks and the number of wins summarize the classification performance over all data sets. A good technique yields a low average rank and a large number of wins. The AL technique with RWM kernel and 4DS selection strategy performs best on 16 of the 20 data sets (highest number of wins) and achieves the smallest average rank. The critical difference (CD) plot of the Nemenyi test is shown in Figure~\ref{fig:CDplot}. As the differences of the average ranks between the active learner for SVM with RWM kernel and 4DS selection strategy and the other two paradigms are greater than the CD we can state that the former performs significantly better than the latter mentioned ones.
	
\begin{table*}[htb]
	\caption{Classification accuracy (in \%) on test data (5-fold cross validation results) for AL of SVM with RBF kernel and US selection strategy, $\text{CMM}_{sha}$ and 4DS selection strategy, and SVM with RWM kernel and 4DS selection strategy. The best results are printed in bold face. A good paradigm yields a small rank and a large number of wins.}
	\label{tab:Accuracies}
	\centering
	\vspace{0.5em}
	\scriptsize{
		\begin{tabular}{l c c c }
			\toprule
			Data Set & RBF kernel with US  & $\text{CMM}_{sha}$ with 4DS  & RWM kernel with 4DS  \\
\midrule
Australian & 84.06 & 81.01 & \textbf{85.65} \\
Clouds & 77.20 & \textbf{89.30} & 88.92 \\
Concentric & 99.52 & 97.32 & \textbf{99.64} \\
CreditA & 84.35 & 77.97 & \textbf{85.65} \\
CreditG & 71.20 & 69.60 & \textbf{72.40} \\
Ecoli & \textbf{85.73} & 80.30 & 85.15 \\
Glass & 65.89 & 56.08 & \textbf{71.01} \\
Heart & 82.96 & 81.11 & \textbf{84.81} \\
Iris & 96.67 & 84.00 & \textbf{98.00} \\
Page Blocks & 93.06 & 94.37 & \textbf{94.52} \\
Phoneme & 80.50 & 79.22 & \textbf{80.66} \\
Pima & \textbf{76.04} & 70.18 & 75.00 \\
Ripley & 88.96 & 90.24 & \textbf{90.40} \\
Satimage & 75.39 & 83.92 & \textbf{86.33} \\
Seeds & 91.43 & 92.86 & \textbf{97.62} \\
Two Moons & 95.50 & 99.99 & \textbf{100.00} \\
Vehicle & \textbf{80.02} & 61.11 & 76.84 \\
Vowel & 77.98 & 88.89 & \textbf{93.23} \\
Wine & 97.21 & 96.06 & \textbf{98.32} \\
Yeast & 56.94 & 44.47 & \textbf{58.08} \\[0.5em]
Mean & 83.03 & 80.90 & 86.11 \\[0.5em]
Rank & 2.200 & 2.600 & 1.200 \\
Wins & 3.0 & 1.0 & 16.0 \\
\bottomrule
		\end{tabular}
	}
\end{table*}

%

\begin{table*}[!ht]
	\caption{Summary of performance regarding additional evaluation criteria. Larger, positive mean AULC values, smaller mean DUR values and CDM values close to zero are indicators for a good AL approach. }
	\label{tab:PerfromanceMeasures}
	\centering
	\vspace{0.5em}
	\scriptsize{
		\begin{tabular}{l c c c  c  c  c }
			 \toprule 
			 \multirow{2}{*}{Active Learner}& \multicolumn{2}{c}{AULC} & \multicolumn{2}{c}{DUR} & \multicolumn{2}{c}{CDM} \\
			 \cmidrule(lr){2-3} \cmidrule(lr){4-5} \cmidrule(lr){6-7}
			 & Mean & Wins & Mean & Wins & Mean & Wins \\
			 \midrule
			 RBF kernel with US  & 0.000 & 7.0 & 1.000 & 6.0 & 0.080 & 4.0 \\[0.25em]
			 $\text{CMM}_{sha}$ with 4DS & -1.407 & 2.5 & 1.970 & 1.5 & 0.058 & 3.5 \\[0.25em]
			 RWM kernel with 4DS  & 2.978  & 10.5 & 0.947 & 12.5 & 0.048 & 12.5 \\
			 \bottomrule
		\end{tabular}
	}
\end{table*}

Furthermore, three additional evaluation measures are used to assess our results numerically: (1) the area under the learning curve (AULC)~\cite{CKS06}, (2) the data utilization rate (DUR)~\cite{CKS06}, and (3) the class distribution match (CDM)~\cite{RS13}. The evaluation results are summarized in Table \ref{tab:PerfromanceMeasures}. A good active learner achieves a large, positive AULC value, a DUR value less than one, a CDM value close to zero, and, of course, a large number of wins for each of the evaluation measures. Table \ref{tab:PerfromanceMeasures} shows that the AL technique with RWM kernel and 4DS selection strategy outperforms the other two paradigms regarding all evaluation criteria.


\section{Challenges for Future Research on Collaborative Active Learning -- \textit{What are the unanswered questions that we will address?}}
\label{sec:Challenges}

Up to now, we discussed the state of the art and, in particular, our own efforts to improve the state of the art. In the preceding section we have shown that significant achievements were made. Now, the curtain falls, and the stage is set for the next scene: \textit{collaborative active learning (CAL)}. In our future work we will answer many questions, most of which caused by the harsh limitations sketched in Section \ref{sec:Introduction}. To give some examples for questions:
%
%
Is it possible to train a classifier actively with labels that are subject to uncertainty? Whom do we ask for labels, if there is a ``pool'' of experts available? Do we query label information from more than one expert? How do we exploit the various, possibly contradictory label information? Is it cheaper or faster, to query for label information for a process (e.g., modeled by a rule premise for which a conclusion has to be found) instead for a batch of samples? 
How can we give feedback to the experts and how can the experts in turn learn from the active learner? How do we determine if we reached a saturation point (e.g., more label information will not increase the performance)? 

\subsection{Challenge 1: Uncertain Oracles}
In a first step, we address the very obvious fact that oracles are not always right. In principal, labels are subject to uncertainty. Here, the meaning of the term uncertainty is adopted from \cite{MS97}. 
That is, ``uncertain'' is a generic term to address various aspects such as ``unlikely'', ``doubtful'', ``implausible'', ``unreliable'', ``imprecise'', ``inconsistent'', or ``vague''.

In real-world applications the labels may come from various sources, often but not always humans. Therefore, a new problem arises: The labels are subject to uncertainty for different reasons. For example, the performance of human annotators depends on many factors: e.g., expertise/experience, concentration/distraction, boredom/disinterest, fatigue level, etc. Furthermore, some samples are difficult for both experts and machines to label (e.g., samples near the decision boundary). Results of real experiments or simulations may be influenced, too: There may be stochasticity which is inherent to a certain process, sensor noise, transmission errors, etc., just to mention a few. Thus, we face many questions: How can we make use of uncertain oracles (annotators that can be erroneous)? How do we decide whether an already queried sample has to be labeled again? How do we deal with noisy experts whose quality varies over time (e.g., they gather experience with the task, they get fatigued)? How does remuneration influence the labeling quality of a noisy expert (e.g., if they are payed better, they are more accurate)? How can we decide whether the expert is erroneous or an observed process itself is nondeterministic? 

As a starting point, we may assume that the ``expertise of an expert'' (i.e., the degree of uncertainty of an oracle) is time-invariant and global in the sense that it does not depend on certain classes, certain regions of the input space of the model to be learned (e.g., a classifier), etc. Then, we may ask experts for either (1) one class label with a degree of confidence, (2) membership probabilities for each class (with or without confidence labels), (3) lower bounds for membership probabilities (cf.\ \cite{AS09}),
(4) a difficulty estimate for a data object that is labeled,
(5) relative difficulty estimates for two data objects (``easier'' or ``more difficult'' to label), etc. Then, we have to define appropriate ways to model that uncertainty (e.g., second-order distributions over parameters of class distributions in a probabilistic framework) and to consider it in selection strategies (e.g., with additional criteria) and for the training of a classifier (e.g., with gradual labels). 
Different query strategies based on density power divergence~\cite{BSP11} (such as $\beta$-divergence and $\gamma$-divergence) for PAL are proposed in \cite{SUK+13}, under the assumption that there is a binary classification problem to solve and wrong labels are induced by careless mistakes or failures of experimental instruments. Thus, the noise is assumed to be uniformly-distributed over the samples space.


\subsection{Challenge 2: Multiple Uncertain Oracles}
In a second step, we address situations where several, individually uncertain oracles (e.g., several human experts with different degree of expertise) contribute their knowledge to an AL process. Thus, AL will now rely on the collective intelligence of a group of oracles. We see this step as a first important step towards collaborative active learning.

\begin{figure}[htb]
	\centering
	\input{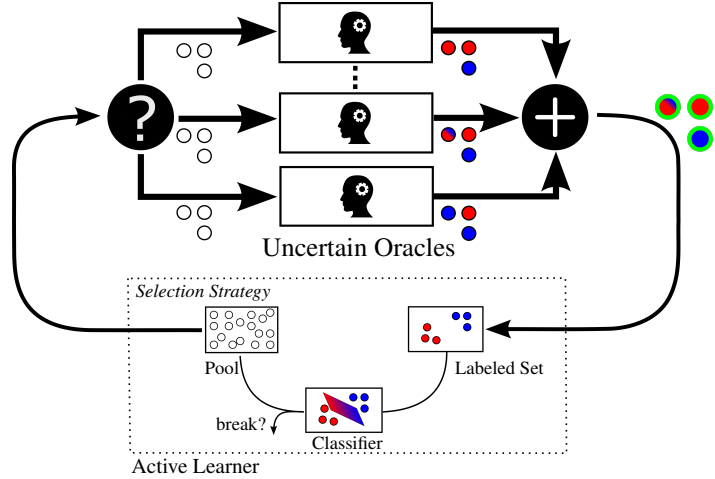}
	\caption{Learning cycle of PAL with multiple uncertain experts.}\label{fig:CALloop}
\end{figure}

In various applications, different, uncertain oracles may contribute labels to an AL process (cf.\ Figure \ref{fig:CALloop}). These experts may not only have different degrees of expertise. They also may have more or less expertise for different parts of the problem that has to be solved, e.g., for different classes that have to be recognized, for different regions of the input space, for different dimensions of the input space (attributes), etc. Also, experts may learn from others and improve over time, for instance. Now, we face many new questions: How can we recognize and model the expertise of humans? How can we decide whom to ask next and how can we merge the uncertain label information (cf.\ also challenge 1)? How can such exploration and exploitation phases be interwoven? Can we identify groups of experts that should cooperate in a labeling process?
How to proceed if experts are only available on a part time basis?

As a starting point, we may initially assume that the ``expertise of an expert'' is known. In principal, we are convinced that generative, probabilistic models can be taken to model the individual knowledge of experts and not only the ``global'' knowledge of the active learner. Uncertainty may again be captured with second-order approaches. New selection strategies must then not only choose samples, but also oracles. If the expertise of an oracle is not known, it must be stated either by asking for difficulty or confidence estimates or by comparing it to the knowledge of others (e.g., by asking an expert who has to be assessed questions with already known answers). In order to explore solutions to challenge 2, we are also confronted with the problem of simulation: We have to simulate several uncertain oracles with the different characteristics mentioned above.

\subsection{Challenge 3: Alternative Query Types}\label{subsec:AlternativeQueryTypes} 

By exploring and modeling the knowledge of oracles as sketched above, the costs of AL would increase substantially. In the other hand, we might ask oracles such as human experts for more abstract knowledge with the goal to reduce the number of queries this way.

In many applications, active learners could ask for more ``valuable'' knowledge. Examples are conclusions that a human expert gives for a presented rule premise, or correlations between different features or features and classes that an expert provides in order to identify important or redundant features. Questions that arise in this context are: Which questions can be asked? How can we provide (i.e., visualize, for instance) the required information to the expert? How can we combine different kinds of expert statements, e.g., about samples, rules, relations between features, etc? How can we use this information to initialize the models that are trained or to restrict the model capabilities in an appropriate way (e.g., if features are known not to be correlated?
                                                                                           
\begin{figure}[htb]
	\centering
	\includegraphics[scale=1.2]{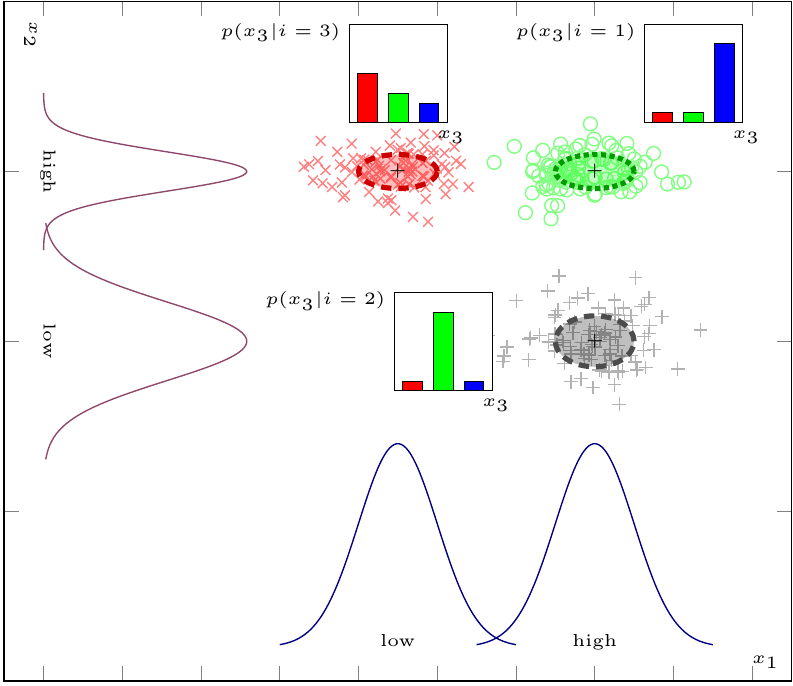}
	\caption{Asking for conclusions of rule premises.}\label{fig:RuleQuery}
\end{figure}

As a starting point, we could investigate the case of annotating rule premises with conclusions. To stay in a probabilistic framework we could obtain user-readable rule premises by marginalization of density functions from a generative process model. Figure \ref{fig:RuleQuery} gives an example for a density model consisting of three components in a three dimensional input space.
The first two dimensions $x_1$ and $x_2$ are continuous and, thus, modeled by bivariate Gaussians whose centers are described by larger crosses (+).
The ellipses are level curves (surfaces of constant density) with shapes defined by the covariance matrices of the Gaussians.
Here, due to the diagonality of the covariance matrices these ellipses are axes-oriented and their projection onto the axes is also shown.
The third dimension $x_3$ is categorical with categories A (red), B (green), and C (blue).
The distributions of $x_3$ are illustrated by the histograms next to every component.
Here, only categories with a probability strictly greater than the average are considered in rules in order to simplify the resulting rules. The components modeling sets of circles (green) and crosses (red) are already labeled, resulting in two rules for the components $i = 1$ and $i = 3$:
\begin{quote}
  \begin{minipage}{\textwidth}
    if $x_1$ is \textit{low} and $x_2$ is \textit{high} and $x_3$ is A or $x_3$ is B
    \hspace*{0.0cm} then class = red,\\
    if $x_1$ is \textit{high} and $x_2$ is \textit{high} and $x_3$ is C
    \hspace*{0.0cm} then class = green.
  \end{minipage}
\end{quote}
Now, the active learner presents the following rule premise and asks for a conclusion in form of a class assignment:
\begin{quote}
  \begin{minipage}{\textwidth}
     $x_1$ is \textit{high} and $x_2$ is \textit{low} and $x_3$ is B.
  \end{minipage}
\end{quote}
This information could then be used to (re-)train a classifier, e.g., in a transductive learning step.

\subsection{Challenge 4: True Collaboration of Human Experts in AL}\label{subsec:TrueCAL} 

In a fourth step we could pave the way for a true collaboration of human experts in AL, which will essentially be based on the capability of humans to learn and the ability of the active learner to provide appropriate feedback to the humans to enable them to learn. Then, the new technique actually deserves to be called CAL.

In particular applications, experts would be interested in getting feedback from an active learner, in improving their own knowledge, and sharing their expertise with others. As an important requirement, the active learner must be able to give feedback to the humans and asking for comments on such feedback. Some possible kinds of interactions with humans are (cf.\ also \cite{HS07}):
\begin{quote}
  \begin{minipage}{\textwidth}
	  The following rule appears to be very certain because ... !\\
    The following rule is in conflict with your knowledge because ... !\\
		Other experts are much less uncertain concerning the following rule than you are ... !\\
		Can you confirm the following rule ... ?\\
		Can you confirm that the following two features are not correlated ... ?\\
		Can you confirm that the following feature is very important ... ?\\
		Can you provide additional samples for the following regions of the input space of the classifier ... ?
  \end{minipage}
\end{quote}
Solutions to this challenge (which is based on appropriate solutions for the preceding three challenges) will open the door for online AL (cf.\ SAL mentioned in Section \ref{sec:Foundations}). Some of the many new questions that have to be answered are: How can we deal with time-invariant knowledge of oracles? Which information should be provided and how (e.g., with/without certainty estimates, restriction to ``crisp'' rules or not)? How must we adapt the active learner and the selection strategies? In particular, a compromise has to be found between modeling capabilities on the one hand and the abilities of humans to actually understand readable rules on the other.

As a starting point, we may stay within our probabilistic framework, consider the individual knowledge of humans (challenge 2) and present samples and rules (obtained by marginalization from density models to make them human-readable as sketched above, challenge 3) with fused statements (labels or conclusions) and certainty estimates. Then, the time-variance of human knowledge must be considered by extending the solutions from challenge 2. Again, the evaluation of any proposed techniques will be a challenge by itself.


\subsection{Challenge 5: Complex Cost Schemes} 

In many real-world applications obtaining class information may be possible at different costs, e.g., some class information is more expensive than other or the labeling costs depend on the location of the sample in the input space. This already applies to a ``conventional'' AL setting without the many ideas discussed in challenges 1 -- 4. In a CAL setting, considering complex cost schemes is even more important.

For CAL applications, we must consider costs that depend on
\begin{enumerate}
	\item \textit{samples with their classes:} As mentioned above, labeling costs may depend on the class (e.g., some kinds of error classes in an industrial production process may be more difficult to detect than others) or on the location of the sample in the input space (e.g., samples close to the decision boundary require higher temporal effort), for instance.
	\item \textit{query types:} It is obvious that different labeling costs have to be foreseen for samples (with or without certainty estimates) or for more complex queries such as rule premises. The cost schemes have to be even more detailed in a CAL setting with feedback to the humans (e.g., with queries such as ``Can you confirm that ...?'').
	\item \textit{oracles (experts):} The costs of humans may depend on their expertise, their temporal effort, their availability (e.g., working hour may be modeled with finite costs, otherwise costs are infinite), etc.
\end{enumerate}
In principle, all these costs may change over time, too. The basic questions in this context are: How can a cost scheme be defined? How must the selection strategies be adapted?

As a starting point, we suggest to choose the first point from the list above and investigate solutions in a ``classical'' AL setting. Then, the most important case for CAL must be addressed, second point. After successfully extending the cost schemes to address the first two tasks we advance to apply CAL with human experts, third point.

\subsection{Further Challenges}

Some other important challenges must be addressed as well or they will be subject to future research. Apart from these challenges we still face the already discussed requirements such as ``parameter-free'' AL or self-adaptation of selection strategies. 

\begin{enumerate}
\item{\textit{Stopping Criterion:}} Currently, the stopping criterion in real-world applications is based on economic factors, e.g., the learner queries samples as long as the budget allows it. The challenge consists in knowing when to stop querying for labels. One possibility may be to determine the point at which the cost of querying more labels is higher than costs for misclassification. Another possibility is to determine when the learner is at least as good as the group of annotators. 
For such a ``self-stopping criterion'', the active learner should be able to assess its own performance. 

\item{\textit{Performance Assessment:}} In AL, the performance of an active learner must be assessed by means of several criteria to capture effectiveness and efficiency of AL. For this purpose, we used a ranked performance measure, a data utilization measure, the area under the learning curve, and a class distribution measure in our preliminary work (see, e.g., \cite{RS13,RCS14}). CAL requires additional measures, e.g., to assess the various learning costs or to evaluate the learning progress of human experts.

\item{\textit{Dynamic Environment:}} Above, we have sketched CAL which takes place in a time-variant environment in the sense that the knowledge of experts improves over time. But, the observed and modeled processes could be time-variant, too. That is, these processes may change slightly (e.g., due to increased wear), become obsolete or new processes corresponding to known or to new, previously unknown, classes may arise during the application of the model. Then, a major challenge consists in developing online AL / CAL techniques that cope with such effects. Mixtures of PAL and SAL techniques would be needed.
\end{enumerate}


\section{Summary and Outlook}\label{sec:SummaryOutlook}

In this article, we have sketched our vision of collaborative active learning which will certainly be discussed in more detail in the near future. In the novel field of CAL, we would like to concentrate on developing classifiers that take class information uncertainty into consideration, identifying the annotators' level of expertise,
making use of different levels of expertise and fusing possibly contradicting knowledge,
labeling abstract knowledge, and
 improving the expertise of the experts. 
In the envisioned CAL system, human domain experts should benefit from sharing their knowledge in the group. They should receive feedback which will improve their own level of expertise.

In principal, many application areas could benefit from CAL techniques. We can distinguish two possible basic cooperation scenarios: First, scenarios involving specialists (e.g., industrial experts) and, second, scenarios involving non-experts (e.g., crowd-sourcing).

In the former scenario, the number of humans will be lower, the humans are motivated, their expertise will be easier to capture, they collaborate over longer time periods, etc. 
Typical industrial problems are, for example, product quality control (e.g., deflectometry, classification of errors on silicon wafers or mirrors, analysis of sewing or garments in clothing industry, etc.), fault detection in technical and other systems (e.g., analysis of fault memory entries in control units of cars, analysis of different kinds of errors in cyber-physical systems, etc.), planing of product development processes (e.g., in drug design), or fraud detection and surveillance (e.g., credit card fraud, detection of tax evasion, intrusion detection, or video surveillance). This scenario may be called \textit{Dedicated CAL}.

In the latter scenario, we face larger, open groups of people that will be available for shorter time spans. 
Typical crowd-sourcing applications will address problems where queries (samples or rules) can easily be understood and assessed by many people including ``non-experts'' and, thus, be based on video, audio, text or image data. CAL may even be a core component of recommender systems, e.g., to suggest television programs. This scenario may be called \textit{Opportunistic CAL}, as the active learner hay to make the most of the current situation.

\todo{BS@AC: Literaturstellen sehr uneinheitlich, fehlende Informationen, evtl teilw. falscher Bib-Typ.}

\biboptions{sort&compress}
\bibliographystyle{elsart-num-sort.bst}
\bibliography{bib/bibliography}

\listoftodos

\end{document}